\documentclass{llncs}
\usepackage{graphicx}
\usepackage{amsmath,amssymb} 
\usepackage{color}
\usepackage[width=122mm,left=12mm,paperwidth=146mm,height=193mm,top=12mm,paperheight=217mm]{geometry}

\usepackage[nocompress]{cite} 
\usepackage{csquotes}
\usepackage{times}
\usepackage{epsfig}
\usepackage{siunitx}
\usepackage{bm}
\pdfoutput=1

\begin{document}
\mainmatter
\def\ECCV18SubNumber{2551}  

\title{Saliency Preservation in Low-Resolution Grayscale Images} 



\author{
	Shivanthan A.C. Yohanandan (shivanthan.yohanandan@rmit.edu.au), Adrian G. Dyer, Dacheng Tao, and Andy Song
}

\institute{RMIT University, Melbourne, Australia}

\maketitle

\begin{abstract}
Visual salience detection originated over 500 million years ago and is one of nature's most efficient mechanisms. In contrast, many state-of-the-art computational saliency models are complex and inefficient. Most saliency models process high-resolution color (HC) images; however, insights into the evolutionary origins of visual salience detection suggest that achromatic low-resolution vision is essential to its speed and efficiency. Previous studies showed that low-resolution color and high-resolution grayscale images preserve saliency information. However, to our knowledge, no one has investigated whether saliency is preserved in low-resolution grayscale (LG) images. In this study, we explain the biological and computational motivation for LG, and show, through a range of human eye-tracking and computational modeling experiments, that saliency information is preserved in LG images. Moreover, we show that using LG images leads to significant speedups in model training and detection times and conclude by proposing LG images for fast and efficient salience detection.


\end{abstract}

\section{Introduction}

Visual scenes often contain more items than can be processed concurrently due to the visual system's limited processing capacity \cite{mcmains_visual_2009}. Visual salience (or attention) detection is a cognitive mechanism that efficiently deals with this capacity limitation by selecting relevant or salient information, while ignoring irrelevant information \cite{mcmains_visual_2009}. Concretely, visual salience refers to conspicuous regions or objects that stand-out or ‘pop-out’ in the visual field or an image, mainly due to apparent differences to their surroundings \cite{borji_what_2013}. Salience detection is a fundamental vision mechanism present in many sighted organisms. Even insects, despite having significantly smaller brains and dissimilar eyes to vertebrates, can detect salient stimuli in their visual field \cite{morawetz_visual_2012,avargues-weber_forest_2015,morawetz_blue_2013}.

Visual salience detection can be crudely divided into bottom-up and top-down mechanisms. Bottom-up salience is stimulus and feature-driven, and responsible for automatic, involuntary rapid shifts in attention and gaze. In contrast, top-down salience is experience-based and varies between individuals \cite{hou_visual_2013}. In computer vision, most studies focus on bottom-up models because they are likely to succeed or be effective in real circumstances. Top-down models are biased, require prior knowledge about the visual content, and are sluggish at best \cite{hou_visual_2013,juola_theories_2016,ramkumar_modeling_2015}.

Recently, deep neural networks have achieved state-of-the-art performance on various saliency benchmarks \cite{kummerer_deepgaze_2016,huang_salicon:_2015,kruthiventi_deepfix:_2017,cornia_predicting_2016}. Nevertheless, this success comes at high computational costs \cite{rajankar_international_nodate,wang_deep_2017}. Training and running these networks is time- and resource-intensive, which is not easily scalable to resource-limited devices \cite{rajankar_international_nodate}. Processing high-resolution or stacked multi-resolution color images is resource-intensive and contributes to this limitation \cite{vo_processing_2016}. In contrast, natural visual salience detection proves to be much more efficient. A deeper understanding of the evolutionary origins of visual salience detection suggests that bottom-up saliency is computed from achromatic low-resolution information \cite{lamb_evolution_1995}.

Previous studies have shown that low-resolution \textit{color} (LC) \cite{judd_fixations_2011,shen_learning_2014,ho-phuoc_when_2012} and \textit{high-resolution} grayscale (HG) \cite{hamel_contribution_2014,hamel_contribution_2016,hamel_does_2015,frey_whats_2008,baddeley_high_2006,dorr_colour_2010} images preserve saliency information, yet are significantly more computationally attractive than high-resolution color (HC) images. Low-resolution grayscale (LG) images are even more computationally attractive, compared to LC and HG images. Nevertheless, to our knowledge, no one has investigated whether saliency information is preserved in LG images. In this study, we therefore investigate saliency preservation in LG images, and present the following three contributions: (1) linking low-resolution grayscale information with the bio-inspired evolutionary origins of visual saliency, (2) assessing the preservation of saliency information in low-resolution grayscale images, and (3) proposing low-resolution grayscale images for fast and efficient saliency detection. Therefore, based on a deeper understanding of the evolutionary origins of visual saliency, together with knowledge gained from studies investigating salience preservation in LC and HG images, we hypothesize that saliency information is well-preserved in LG images.

\section{Related work}

\subsection{Fixations on low-resolution images}

Judd \textit{et al.} \cite{judd_fixations_2011} investigated how well fixations on LC images predict fixations on the same images in HC. They found that fixations on LC images ($76 \times 64$ pixels) can predict fixations on HC images ($610 \times 512$ pixels) quite well (AUC-Judd \cite{judd_fixations_2011} $> 0.85$). However, they did not investigate the HC fixation-predictability of LG images, nor did they mention any biological plausibility for deciding to investigate fixations in LC images. Nevertheless, they concluded that working with fixations on LC instead of HC images could be perceptually adequate and computationally attractive, which is part of our motivation for pursuing this study.

\subsection{Multi-resolution approaches}

Deep artificial neural networks are not inherently scale-invariant \cite{xu_scale-invariant_2014}. Therefore, multi-resolution models are often used to capture saliency at different scales. Advani \textit{et al.} \cite{advani_multi-resolution_2013} presented a multi-resolution framework for detecting visual salience where resolution degrades further away from the point of fixation represented as a three-level architecture: a central high-resolution fovea ($960 \times 960$ pixels), a mid-resolution filter ($640 \times 640$ pixels), and a low-resolution region ($480 \times 480$ pixels). They found significant computational benefits using this model, but only investigated color images and ignored the achromaticity of peripheral vision.

Shen \textit{et al.} \cite{shen_learning_2014} went a step further and modeled the visual acuity of the parafovea and periphery as a stack of multi-scale inputs. They extracted multi-resolution image patches in multiple visual acuity on the same image from fixation targets and non-target locations based on the “sunflower” model of retina \cite{lindeberg_foveal_1994,koenderink_visual_1978,romeny_front-end_2008}. However, despite finding comparable performance to higher-resolution models, they too only investigated color images, and overlooked the fact that the parafovea and periphery predominantly processes achromatic information \cite{lamb_evolution_1995}. Furthermore, multi-scale models need to process and train on the same image multiples times at different resolutions, which is computationally unattractive. Therefore, the ideal input image has the lowest resolution and smallest color space that preserves saliency.

\subsection{Fixations in grayscale}

Hamel \textit{et al.} \cite{hamel_does_2015} investigated the role of color in visual attention by comparing eye movements across different participants viewing color and grayscale videos. They found color to only have a modest effect in predicting salience. However, they only investigated high-resolution images, leaving the influence of color in low-resolution images a gap for us to fill.

Yang \textit{et al.} \cite{yang_color_2010} also investigated whether saliency information is preserved in grayscale images using a novel minimization function. They showed that saliency is well-preserved in grayscale images of the same resolution, but did not extend their investigation to lower resolutions, which our study aims to do.

\section{Evolutionary origin of visual saliency}
\begin{figure}
\begin{center}
\includegraphics[width=0.6\linewidth]{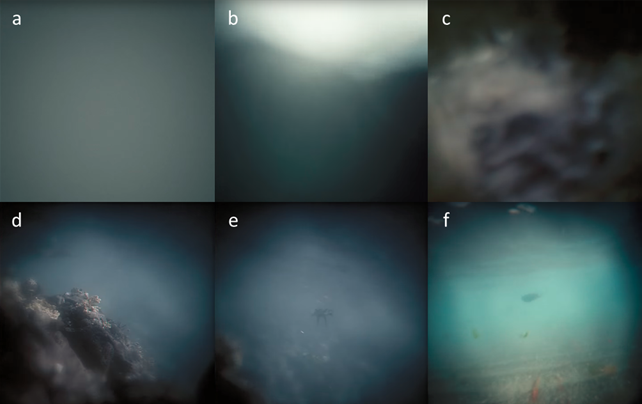}
\end{center}
	\caption{Hypothetical stages of the evolution of vertebrate vision. This figure panel shows a series of photographic reconstructions of how the vertebrate eye is hypothesized to have evolved, and what that vision hypothetically looked like from an animal's point of view. Static images adapted from \textit{Cosmos: A Spacetime Odyssey (Some of the Things that Molecules do)} \cite{pope_cosmos:_2014}.}
\label{fig:originOfVision}
\end{figure}

In the beginning, life was blind. Then, around 600 million years ago, the first eyes discriminated night and day (Figure \ref{fig:originOfVision}(a)) \cite{nilsson_evolution_2009}. Light-source localization followed a few million years later (Figure \ref{fig:originOfVision}(b)), heralding eyes capable of distinguishing light from shadow, thus crudely making-out objects in their vicinity (Figure \ref{fig:originOfVision}(c)), including those to eat, and those that might eat it. This was likely the birth of stimulus-driven, bottom-up visual salience detection -- the mechanism thought to be primarily responsible for the Cambrian explosion \cite{lamb_evolution_1995}. Later, things became a little clearer. The eye's opening contracted to a pinhole covered by a protective transparent membrane, allowing just enough light to paint a dim image on the sensitive inner surface of the eye \cite{potter_detecting_2014}. Then came focus-sharpening lenses (Figure \ref{fig:originOfVision}(d)), foveated central vision (Figure \ref{fig:originOfVision}(e)), and finally, color (Figure \ref{fig:originOfVision}(f)). However, despite the arrival of high-acuity chromatic central vision, blurry achromatic peripheral vision dominates over 90\% of our visual field, and is still the primary information source for bottom-up salience detection -- a relic mechanism conserved through evolution in many species because of its apparent speed and efficiency \cite{lamb_evolution_1995}. Furthermore, many sighted animals completely lack chromatic vision, yet are able to rapidly detect obstacles and avoid collisions in complex environments \cite{stojcev_general_2011}.

The ability then of an organism's pupil to rapidly shift foveal gaze to salient regions suggests that it is peripheral vision that points the sharper, high-resolution foveated (sometimes chromatic) vision to investigate objects and regions further. Eye movements align objects with the high-acuity fovea of the retina, making it possible to gather detailed information about the world \cite{potter_detecting_2014}. Therefore, bottom-up visual salience detection is predominantly a peripheral vision information processing task.

\section{Peripheral vision}
\begin{figure}
\begin{center}
\includegraphics[width=\linewidth]{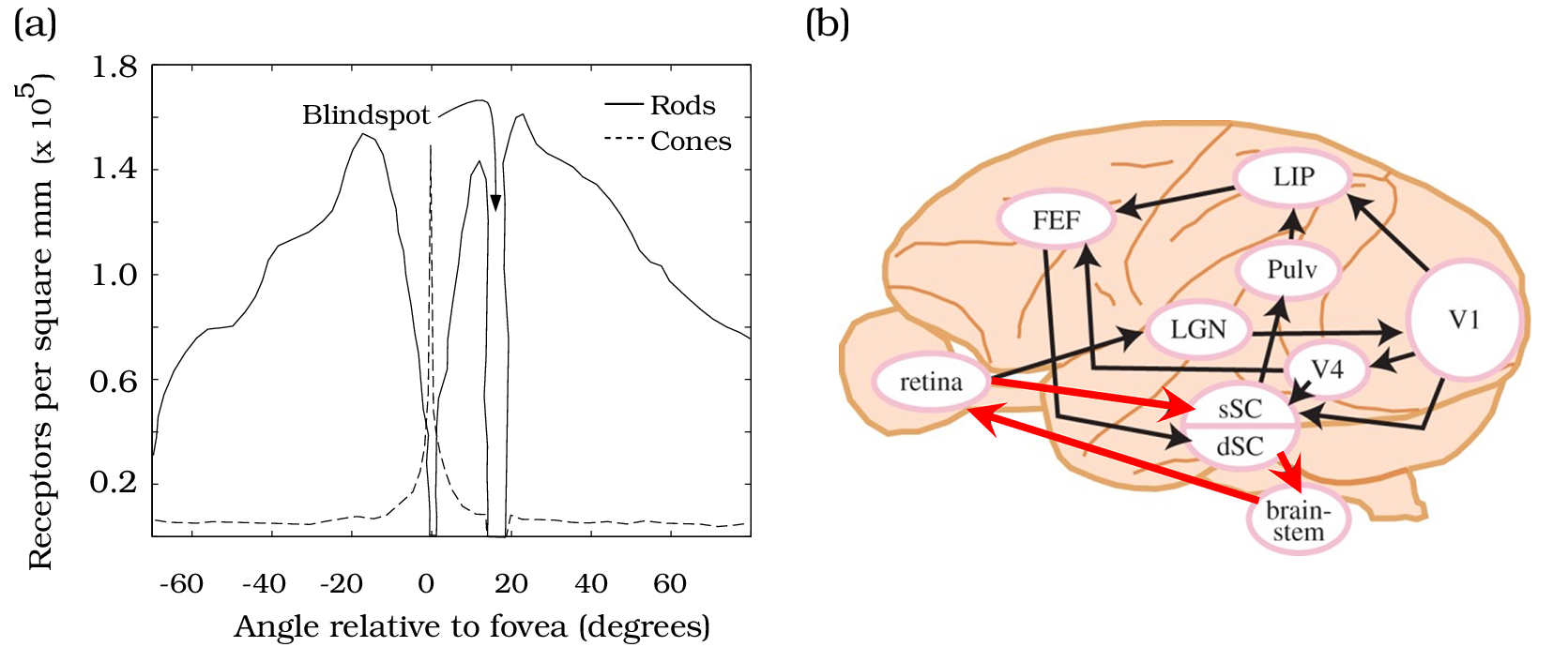}
\end{center}
	\caption{(a) The human retina's distribution of rod and cone photoreceptors is shown in degrees of visual angle relative to the position of the fovea for the left eye. Cones, concentrated in the fovea, encode high-resolution color. Rod photoreceptors distributed outside the fovea encode low-resolution grayscale information \cite{wandell_foundations_1995}. (b) Macaque brain information flow from retinal input to eye movement output. Visual signals from the retina to the cerebral cortex are mediated through the primary visual cortex (V1) and the superior colliculus (sSC and dSC). There is also a shortcut from the superficial (sSC) to the deep (dSC) superior colliculus, which then sends outputs directly to the brainstem oculomotor nuclei, resulting in rapid saccades (red pathway) \cite{veale_how_2017}.}
\label{fig:rodconedistbrainpaths}
\end{figure}

A key to the speed and efficiency in bottom-up salience detection lies in the distribution of rod and cone photoreceptor cells in the human retina (Figure \ref{fig:rodconedistbrainpaths}(a)), and the information processing pipeline of typical vertebrate peripheral vision (Figure \ref{fig:rodconedistbrainpaths}(b)). Rod cells primarily encode achromatic luminance (brightness) information, and have a higher distribution outside the fovea. In contrast, cone cells encode chrominance (color), and are concentrated in the fovea (center of the retina) \cite{wandell_foundations_1995}. Moreover, multiple rod cells converge to and activate a single retinal ganglion neuron, which is why rod vision has lower spatial resolution compared to information encoded by cones, albeit itself having a high peripheral distribution. In contrast, each cone activates multiple ganglion neurons, resulting in higher acuity vision \cite{okawa_optimization_2007}. Therefore, afferent ganglion neurons, not photoreceptors, from the retina determine the perceived image resolution.

The sparse retinal output of peripheral vision enters a structure called the optic tectum (or superior colliculus (SC) in higher-order animals, Figure \ref{fig:rodconedistbrainpaths}(b)). This structure has only recently emerged as a likely candidate for encoding the saliency map –-- a well-known precursor for bottom-up salience detection \cite{veale_how_2017,white_superior_2017,krauzlis_superior_2013}. Furthermore, the SC has direct control of eye muscles. In their study, Veale \textit{et al.} \cite{veale_how_2017} explain that direct retinal input into the SC of a macaque brain can trigger reflex-like saccades via brainstem oculomotor nuclei (red pathway in Figure \ref{fig:rodconedistbrainpaths}(b)). This could explain why bottom-up saliency detection is rapid and reflex-like, which makes sense since it is processing predominantly achromatic information from fewer efferent neurons, compared to foveated vision, which is processed downstream of the SC and in larger complex brain regions, therefore taking longer. This means far fewer neurons enter the SC, which is analogous to a low-resolution grayscale digital image. Therefore, this sparse achromatic peripheral output could be approximated using low-resolution grayscale images in the digital domain.

\section{Approximating peripheral vision}

Leveraging knowledge from Judd \textit{et al.} \cite{judd_fixations_2011} and Hamel \textit{et al.} \cite{hamel_does_2015}, we decided to approximate peripheral vision by first transforming the color space of HC images to 8-bit grayscale (section 5.1), followed by down-sampling the original image height to 64 pixels and width proportionally (section 5.2).

\subsection{Colorimetric grayscale conversion}\label{subsec:color}

Images were first converted from 24-bit sRGB to 8-bit grayscale since it is faster and more efficient to consolidate the three channels before performing subsequent operations, which would otherwise need to be performed thrice (i.e. once per channel). Color to grayscale conversion is a lossy operation, resulting in luminance degradation, which may affect saliency \cite{hamel_contribution_2014}. To avoid such systematic errors, the grayscale conversion must at least preserve the brightness features of the original stimuli (i.e. the luminosity of grayscale pixels must be identical to the original color image). The HC images used in this study are stored in the sRGB (standard Red Green Blue) color space, which also defines a nonlinear transformation (gamma correction) between the luminosity of these primaries and the actual number stored.

To convert the 24-bit sRGB gamma-compressed color model $I_{HC}$ to an 8-bit grayscale representation of its luminance $I_{HG}$, the gamma compression function must first be removed via gamma expansion to transform the image to a linear RGB color space \cite{poynton_rehabilitation_1998}, so that the appropriate weighted sum can be applied to the linear color components $R_{linear}$, $G_{linear}$, $B_{linear}$. For the sRGB color space, gamma expansion is defined as
\begin{equation}
C_{linear}=\begin{cases}\frac{C_{sRGB}}{12.92} & C_{sRGB} \leq 0.04045\\(\frac{C_{sRGB}+0.055}{1.055})^{2.4} & C_{sRGB} > 0.04045\end{cases}
\end{equation}
where $C_{sRGB}$ represents any of the three gamma-compressed sRGB primaries ($R_{sRGB}$, $G_{sRGB}$, and $B_{sRGB}$, each in range $[0,1]$) and $C_{linear}$ is the corresponding linear-intensity value ($R_{linear}$, $G_{linear}$, and $B_{linear}$, also in range $[0,1]$). Then, $I_{HG}$ is calculated as a weighted sum of the three linear-intensity values, which is given by
\begin{equation}
\begin{aligned}
I_{HG}= {} & 0.2126\times R_{linear}+0.7152\times G_{linear}+0.0722\times B_{linear}.
\end{aligned}
\end{equation}
These three coefficients represent the intensity (luminance) perception of a standard observer trichromat human to light of the precise Rec. 709 \cite{poynton_perceptual_2014} additive primary colors that are used in the definition of sRGB.



\subsection{Down-sampling image resolution}

We chose 64 pixels as our low-resolution height since Judd \textit{et al.} \cite{judd_fixations_2011} found this to be the resolution with the best resolution-saliency compromise compared to other resolutions. According to the Nyquist theorem, down-sampling from a higher-resolution image can only be carried out after applying a suitable 2D anti-aliasing filter to prevent aliasing artifacts. To reduce the height of each image down to 64 pixels, we used the same method as Torralba \textit{et al.} \cite{torralba_how_2009}: we first applied a low-pass binomial filter with kernel
\begin{equation}
K=\begin{bmatrix}
1 & 1 & 1 & 1 & 1 \\
1 & 4 & 4 & 4 & 1 \\
1 & 4 & 6 & 4 & 1 \\
1 & 4 & 4 & 4 & 1 \\
1 & 1 & 1 & 1 & 1
\end{bmatrix}
\end{equation}
to $I_{HG}$ and then down-sampled the resulting image using bicubic interpolation by a factor of two, until the desired image height of 64 pixels was reached (corresponding width was maintained based on the original aspect ratio), forming $I_{LG}$. This also had the effect of providing a clear upper bound on the amount of visual information available \cite{torralba_how_2009}.

\section{Experiments}

This section assesses how well saliency information is preserved after transforming HC images to LG images using methods outlined above. Furthermore, it investigates if there are any computational benefits using LG over HC. A fixation map is a two-dimensional spatial record of discrete image locations fixated by an observer, and is collected using an eye-tracker \cite{wooding_fixation_2002}. Previous studies used fixation maps to compare saliency similarity between images \cite{judd_fixations_2011,tavakoli_saliency_2017}. Saliency similarity can also be quantified using fixation-map inter-observer visual congruency (agreement) \cite{tavakoli_saliency_2017}. To that end, we designed and conducted three separate experiments: section 6.1 assesses LG and HC fixation-map similarity; section 6.2 assesses LG vs. HC fixation-map inter-observer congruency; and section 6.3 compares accuracy, training and detection speed performance between saliency models trained on HC and LG data.

\subsection{HC and LG fixation-map similarity}

\textbf{Dataset.} A subset $\bm{I_{HC}}$ of 20 HC images ($1920 \times 1080$ pixels, sRGB) along with the corresponding aggregated eye fixations $\bm{F_{HC}}$ from 18 observers were randomly sampled from the publicly-available CAT2000 benchmark dataset \cite{borji_cat2000:_2015}. This dataset contains 4000 images selected from a wide variety of categories such as \textit{art, cartoons, indoor, jumbled, line drawings, random, satellite, and outdoor}. Overall, this dataset contains 20 different categories with 200 images from each category. Only 20 images were evaluated since the sample size of observers was sufficient to determine the statistical significance of fixation-map similarity. Using methods outlined in section 5, images from $\bm{I_{HC}}$ were first converted to grayscale, then down-sampled to $120 \times 64$ pixels. This resulted in a set of images $\bm{I_{LG}}$ that were a mere 0.12\% of the original size, thus significantly reducing computational costs. For human visualization on the eye-tracker screen, $\bm{I_{LG}}$ images were up-sampled back to their original resolution using the same bicubic interpolation rescaling method outlined in section 5.2.

\textbf{Eye tracking.} Eye fixations $\bm{F_{LG}}$ were collected using a Tobii T60 eye-tracker by allowing 18 consenting participants to free-view each $\bm{I_{LG}}$ image for 3 seconds from a viewing distance of 60 \si{cm}, consistent with the CAT2000 study. Such a viewing duration typically elicits 4-6 fixations from each observer. This is sufficient to highlight a few points of interest per image, and offers a reasonable testing ground for saliency models \cite{bylinskii_what_2016}. Each observer underwent an initial five-point calibration procedure to minimize eye-tracking calibration errors. Every pair of LG/HC images was displayed at least 2 images apart to minimize the effect of priming.

\begin{figure}
\includegraphics[width=\linewidth]{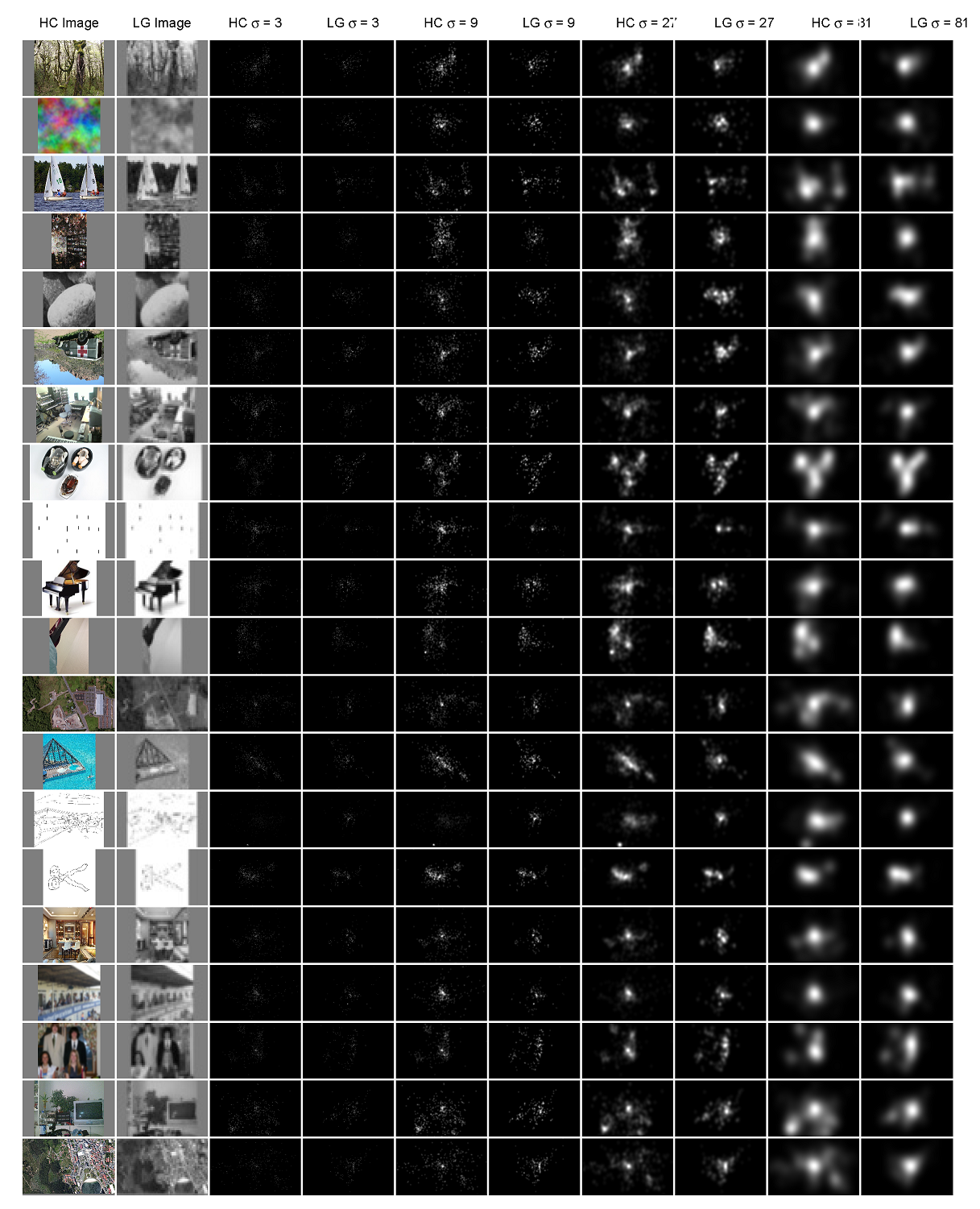}
	\caption{Twenty images from the CAT2000 dataset [46] in high-resolution color (HC) and low-resolution grayscale (LG), and their corresponding
fixation maps (from 18 observers each) as a function of $\sigma$, where $\sigma \in \{3, 9, 27, 81\}$, from Experiment 1 analyses (section 6.1).}
\label{fig:experiment1}
\end{figure}

\textbf{Evaluation metrics.} We compared $\bm{F_{HC}}$ and $\bm{F_{LG}}$ fixation map similarity as a function of six recommended \enquote{gold standard} metrics: Normalized Scanpath Saliency (NSS) \cite{peters_components_2005}, Kullback-Leibler divergence (KL) \cite{liang_top_2015}, Judd Area under ROC Curve (jAUC) \cite{judd_learning_2009}, Shuffled AUC (sAUC) \cite{borji_analysis_2013}, Pearson’s Correlation Coefficient (CC) \cite{le_meur_predicting_2007}, and Similarity or histogram intersection (SIM) \cite{koch_shifts_1985}. These metrics have been used in the past to evaluate fixation map similarity because of their easy interpretability \cite{borji_analysis_2013,bylinskii_what_2016}. We skip explaining these metrics in detail for brevity, and refer readers to the relevant publications.

\begin{figure}
\begin{center}
\includegraphics[width=\linewidth]{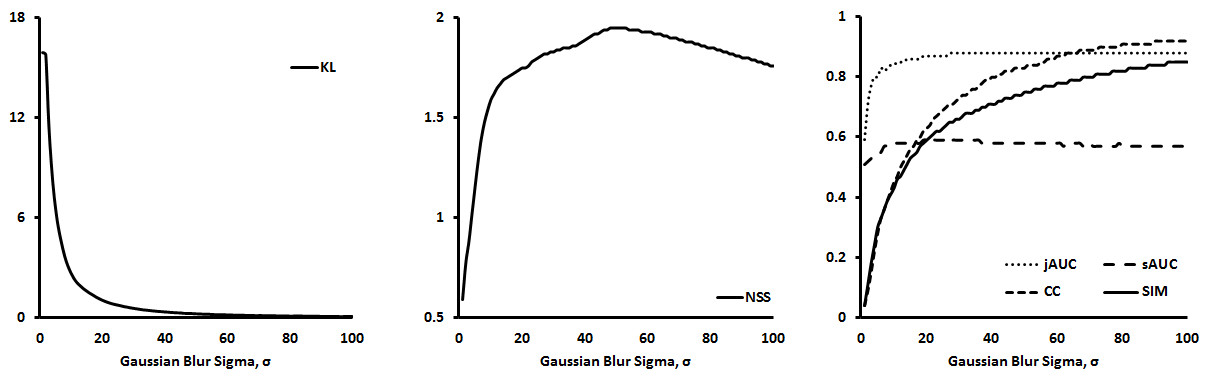}
\end{center}
	\caption{Low-resolution grayscale and high-resolution color fixation-map similarity metrics as a function of the Gaussian blur $\sigma$. Plots represent medians across all participants for all 20 images. Note: NSS $y$--axis range is constrained to min/max and all plots share the $x$--axis.}
\label{fig:metricsplot}
\end{figure}

Discrete fixations from $\bm{F_{HC}}$ and $\bm{F_{LG}}$ are converted into continuous distribution maps $\bm{M_{HC}}$ and $\bm{M_{LG}}$, respectively, by smoothing, which acts as regularization, allowing for uncertainty in the ground truth measurements to be incorporated. A blur value $\sigma$ is required for the Gaussian low-pass filter in the Fourier domain. We follow common practice \cite{bylinskii_what_2016}, and blur each fixation location using a Gaussian with $\sigma$ ranging from 1 to 100, resulting in 100 fixation maps for each HC and LG image per participant. For highly similar fixation maps, all evaluation metrics rise (except KL, which falls) rapidly towards a large maximum as $\sigma \to 100$. Conversely, for highly dissimilar fixation maps, evaluation metrics decrease with an increasing $\sigma$ \cite{engelke_comparative_2013}. We calculated these metrics using MATLAB scripts from \cite{bylinskii_what_2016}, and plot the median across all participants for each metric (Figure \ref{fig:metricsplot}).

\textbf{Results.} From visual inspection (Figure \ref{fig:experiment1}), we can see that increasing $\sigma$ smooths the fixation density map and has the effect of filtering out stray fixations with low inter-observer congruency, leaving behind high-confidence fixations. These results suggest that $\bm{M_{HC}}$ and $\bm{M_{LG}}$ are highly similar, attaining high jAUC (0.88), SIM (0.85) and CC (0.92) as $\sigma \to 100$ (Figure \ref{fig:metricsplot}). Moreover, this result confirms saliency preservation in LG images in terms of fixation map similarity.

\begin{figure*}
\includegraphics[width=\linewidth]{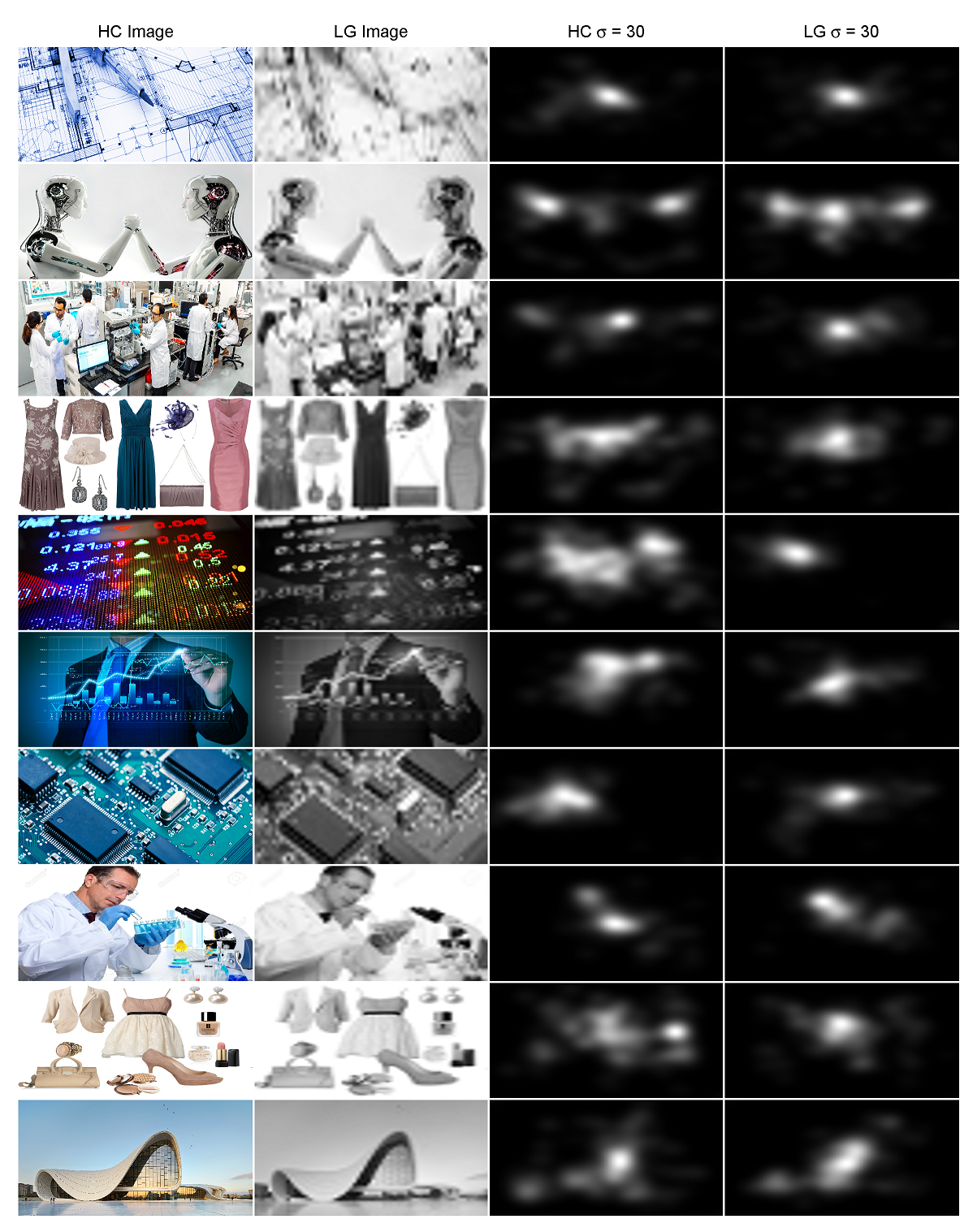}
	\caption{Full set of 10 images in high-resolution color (HC) and low-resolution grayscale (LG), and their corresponding fixation maps (from 35 observers
each) as a function of $\sigma = 30$, from Experiment 2 analyses (section 6.2).}
\label{fig:experiment2}
\end{figure*}

\begin{figure*}
\includegraphics[width=\linewidth]{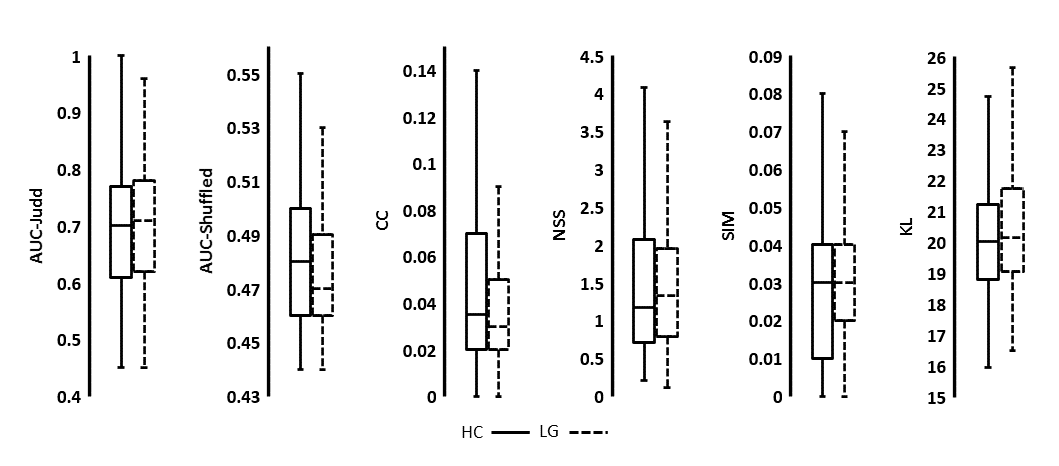}
	\caption{Experiment 2 (section 6.2) boxplots showing high-resolution color (HC) vs. low-resolution grayscale (LG) inter-observer congruency
across 35 observers across 10 images (in HC and LG) as a function of AUC-Judd, AUC-Shuffled, CC, NSS, SIM, and KL. A $\sigma$ of 30
was chosen from Experiment 1 to generate the fixation maps used in this analysis. ANOVA analysis revealed no statistically significant
difference between HC and LG across all 6 metrics. Note: $y$--axes have been cropped and scaled for viewing convenience; boxplot key (from
bottom): minimum, $25^{\text{th}}$ percentile, median, $75^{\text{th}}$ percentile, and maximum.}
\label{fig:congruencyplots}
\end{figure*}

\subsection{HC vs. LG inter-observer consistency}

\textbf{Dataset.} To determine LG and HC inter-observer congruency (agreement), a subset $\bm{I_{HC}}$ of 10 HC ($1280 \times 1024$ pixels, sRGB) images were randomly sampled from the Internet (Google Images) and converted to $120 \times 64$ pixel LG images $\bm{I_{LG}}$ using the same methods described in section 5. As with the previous experiment (section 6.1), 10 images were deemed sufficient to determine statistical significance since the sample size of observers was large. This resulted in images that were only 0.19\% of the original size; once again, significantly reducing computational costs.

\textbf{Eye tracking.} To conduct this analysis, we required separate fixation data from each observer, which was lacking from the CAT2000 dataset's aggregated fixations. To that end, we collected eye-tracking fixation data $\bm{F_{HC}}$ and $\bm{F_{LG}}$ using the same Tobii eye-tracker from 35 consenting observers viewing both sets of $\bm{I_{HC}}$ and $\bm{I_{LG}}$ images, respectively. Standard five-point eye-tracker calibration was performed at the start of each trial for each participant as standard practice. Similar to the previous experiment in §6.1, images were presented for 3 seconds each, and participants were instructed to freely view images, while seated 60 $\si{cm}$ in front of the screen.

\textbf{Evaluation metrics.} We chose a Gaussian blur $\sigma$ of 30, which corresponds to 1 degree of visual angle \cite{bylinskii_what_2016}, generated continuous fixation maps, $\bm{M_{HC}}$ and $\bm{M_{LG}}$, and calculated inter-observer congruency as a function of the same previous set of 6 metrics within the $\bm{F_{HC}}$ and $\bm{F_{LG}}$ sets using the leave-one-out (one-vs-all) method described in \cite{tavakoli_saliency_2017}. We also performed an ANOVA analysis across all co-variates.

\begin{table}
\caption{High-resolution color (HC) vs. low-resolution grayscale (LG) inter-observer consistency results. The values represent medians for each metric across all participants and images. The Gaussian blur $\sigma$ was set to 30 for generating fixation maps.}
\begin{center}
\begin{tabular}{|c|c|c|c|c|c|c|}
\hline
\textbf{} & \textbf{jAUC} & \textbf{sAUC} & \textbf{CC} & \textbf{NSS} & \textbf{SIM} & \textbf{KL} \\ \hline
HC & 0.70 & 0.48 & 0.03 & 1.17 & 0.03 & 20.03 \\
LG & 0.71 & 0.47 & 0.03 & 1.33 & 0.03 & 20.15 \\ \hline
\end{tabular}
\end{center}
\label{tbl:table1}
\end{table}

\textbf{Results.} Table \ref{tbl:table1} and Figure \ref{fig:congruencyplots} show that the LG fixation data does not show a higher dispersion between observers' eye tracking data compared to HG fixations. Furthermore, the ANOVA analysis found no significant difference between HC and LG inter-observer consistency ($p > 0.05$). This result suggests that LG fixation data is as accurate as expected for substituting HC fixation data \cite{tavakoli_saliency_2017}. Moreover, this result further confirms saliency preservation in LG images in terms of fixation map inter-observer congruency.

\subsection{HC vs. LG saliency detection models}

\begin{figure}
\begin{center}
\includegraphics[width=\linewidth]{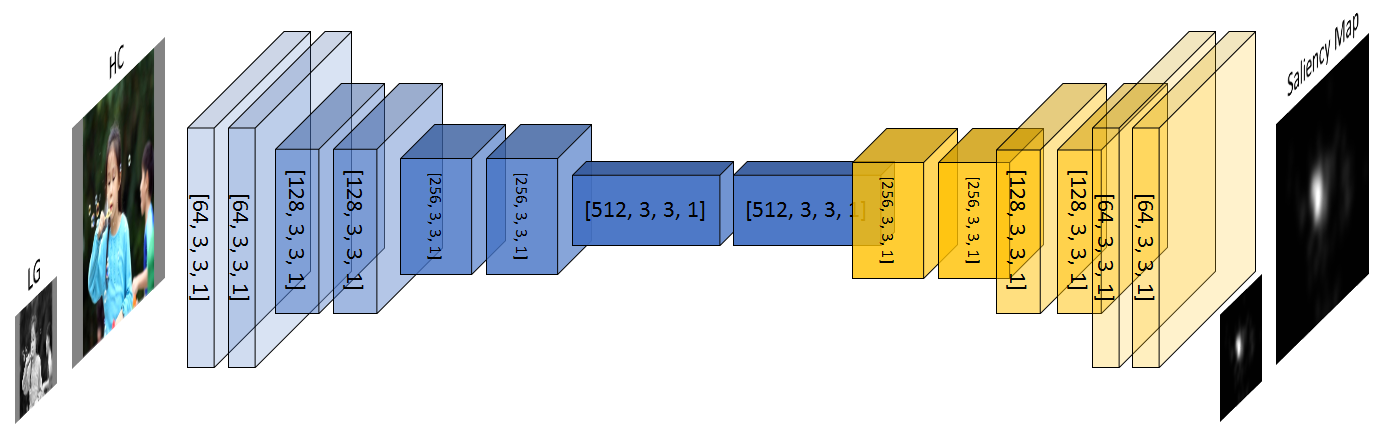}
\end{center}
	\caption{Fully convolutional neural network architecture. The network takes an HC or LG image as input, adopts convolution layers (blue) to transform the image into multidimensional feature representations, then applies a stack of deconvolution layers (orange) for upsampling the extracted coarse features. Finally, a fully convolution layer with a $1 \times 1$ kernel and sigmoid activation function outputs a pixel-wise probability (saliency) map the same size as the input, where larger values correspond to higher saliency. Numbers represent \textbf{[\textit{convolutional filters, kernel width, kernel height, stride}]}.}
\label{fig:fcn}
\end{figure}

\textbf{Model architecture.} Conventional convolutional neural networks (CNNs) used for image classification consists of convolutional layers followed by fully connected layers, which takes an image of fixed spatial size as input and produces a single-dimensional vector indicating the class-probability or category of the input image. For tasks requiring spatial labels, like generating pixel-wise saliency heatmaps, we consider fully convolutional neural networks (FCNs) with deconvolutional layers. This architecture has been previously used for saliency detection in video with enormous success \cite{wang_video_2018}, which is why we used a slightly modified version in our study (Figure \ref{fig:fcn}). It is capable of generating saliency maps the same size as the input image, which was ideal for our experiment since we needed to compare the same model on datasets comprising images of different resolutions and color spaces without needing to change model hyperparameters. To test if HC and LG models have similar accuracy, we kept all other parameters constant and only varied the image resolution and color space during compression. We were only interested in a HC saliency detection model with comparable accuracy and performance to the state-of-the-art so we could show that an LG model can achieve the same performance faster and more efficiently. The model generates a saliency heatmap from a given input, which can then be compared with the ground-truth density map, just as in the above experiments. 

\textbf{Dataset.} The 2000 labeled images from the same CAT2000 saliency benchmark dataset used previously was split into training (1800 images) and validation (200 images) sets. These sets were duplicated and preprocessed to produce four new sets: high-resolution 24-bit color training and validation sets, $\bm{T_{HC}}$ and $\bm{V_{HC}}$, created by downsampling the original resolution to $512 \times 512$ pixels (typical resolution used by many state-of-the-art saliency detection models) using methods described in section 5.2, and low-resolution ($64 \times 64$ pixels) 8-bit grayscale sets, $\bm{T_{LG}}$ and $\bm{V_{LG}}$, generated using methods outlined above.

\textbf{Model training.} The Python Keras API with the TensorFlow framework backend was used to implement and train HC and LG FCN models, $\bm{M_{HC}}$ and $\bm{M_{LG}}$, on the respective $1800$ training images end-to-end and from scratch (i.e. randomized initial weights). Network weights and parameters were initialized by seeding a pseudo-random number generator with the same seed for all training sessions and models to ensure everything else remained constant. The training images were propagated through the FCN in batches of 8 and 64 for $\bm{M_{HC}}$ and $\bm{M_{LG}}$, respectively. Due to the FCN's large parameter space, $\bm{M_{HC}}$ batch size was restricted to $8$ so that the $512 \times 512$ images could be accommodated by the available memory ($12$ GB) and resources. Weights were learned using slow gradient decent (RMSProp) over 100 epochs totaling 180,000 iterations. The base learning rate was set to $0.05$, and decreased by a factor of $10$ after $2000$ iterations. A mean-squared error loss function was implemented to compute loss for gradient descent. An NVIDIA Tesla K80 GPU was used for training and inference. Training time (i.e. the time taken to complete all iterations to completion) for each model was recorded.

\textbf{Evaluation metrics.} $\bm{M_{HC}}$ and $\bm{M_{LG}}$ were tested on their respective held-out validation sets, $\bm{V_{HC}}$ and $\bm{V_{LG}}$. The predicted labels from the models' output were up-sampled to match the original dimensions of the ground truth labels ($1920 \times 1080$ pixels) for a fair accuracy evaluation. Model accuracy was defined as a function of NSS, Judd-AUC, SIM, and CC, described above, and computed using MATLAB code from the MIT saliency benchmark GitHub repository \cite{judd_fixations_2011}. Furthermore, detection time, defined as the average time taken by the model to generate a predicted saliency map based on each of the 200 test images, was also measured for $\bm{M_{HC}}$ and $\bm{M_{LG}}$. Finally, two-tailed paired Student’s \textit{t}-tests were performed between HC and LG result pairs to determine if differences were statistically significant.

\begin{figure}
\begin{center}
\includegraphics[width=0.8\linewidth]{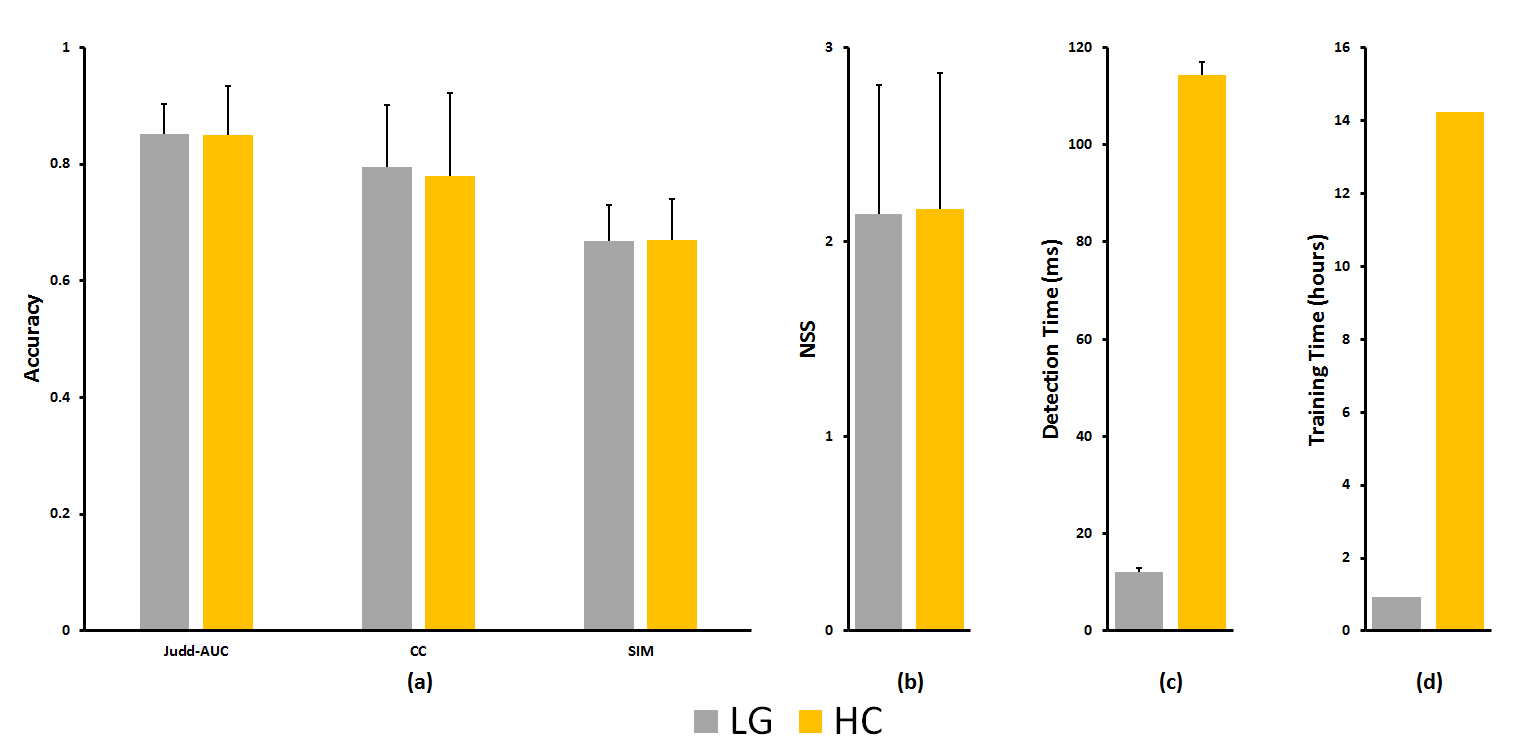}
\end{center}
	\caption{(a) HC and LG model accuracy as a function of Judd-AUC, CC, SIM, and (b) NSS. (c) HC vs. LG training time. (d) HC vs. LG detection time. Bar plots represent means and error bars represent standard deviations across the 200 test results per model.}
\label{fig:model_results}
\end{figure}

\textbf{Results.} Figures \ref{fig:model_results}(a) and \ref{fig:model_results}(b) show no statistically significant difference between $\bm{M_{HC}}$ and $\bm{M_{LG}}$ accuracy across all evaluation metrics ($p > 0.05$). Furthermore, these accuracies are comparable to state-of-the-art models. Therefore, this is further evidence suggesting saliency is well-preserved in LG images. However, Figures \ref{fig:model_results}(c) and \ref{fig:model_results}(d) show a significant difference between $\bm{M_{HC}}$ and $\bm{M_{LG}}$ training and detection times ($p < 0.05$). $\bm{M_{LG}}$ trained more than $14\times$ faster than its HC counterpart, $\bm{M_{HC}}$. Furthermore, $\bm{M_{LG}}$ is capable of generating a predicted saliency map almost $10\times$ faster than $\bm{M_{HC}}$ ($12$ vs. $114$ milliseconds). Considering these significant speedups come at negligible accuracy cost, the implications of using LG images over HC are substantial; thus, the motivation to use LG images in saliency detection should now be more obvious and appealing.

\section{Conclusion}

In this study, we explained and demonstrated the biological and computational motivation for using LG images in salience detection. We learned, through evolutionary insights, that bottom-up visual salience detection is predominantly a peripheral vision mechanism. We also learned that peripheral vision information is primarily achromatic and low-resolution, and can be approximated in the digital domain using a simple LG transformation. Through eye-tracking experiments, we found high similarity between LG and HC fixations. The results of this study also showed no significant difference in inter-observer congruency between LG and HC groups. Additionally, we trained fully convolutional neural networks for saliency detection using LG and HC data from a benchmark dataset and found no significant difference between HC, LG and state-of-the-art model accuracy. However, we found that the LG model required significantly less ($1/14$) training time and is much faster (almost $10\times$) performing detection compared to the same network trained and evaluated on HC images. Therefore, these results confirm our hypothesis that saliency information is preserved in LG images, and we conclude by proposing LG images for fast and efficient saliency detection. Future research will extend this work by investigating the use of LG images in other computer vision tasks, such as object detection, pose tracking and background subtraction, since we have reason to believe that many vision tasks could just as easily be done using peripheral vision and hence, low-resolution grayscale information.

\section{Acknowledgements}

This research was supported by an Australian Postgraduate Award scholarship and the Professor Robert and Josephine Shanks scholarship. The authors wish to thank the eye tracking participants for volunteering their time.

\bibliographystyle{splncs}
\bibliography{ms}
\end{document}